\documentclass[letterpaper, 10 pt, conference]{ieeeconf}  

\IEEEoverridecommandlockouts                              

\usepackage{amsmath}
\usepackage{amssymb}
\usepackage{amsfonts}
\usepackage{mathtools}
\usepackage{bm}
\usepackage{graphicx}
\usepackage{booktabs}
\usepackage{tabularx}
\usepackage{multirow}
\usepackage{flushend}
\usepackage[lined,ruled,linesnumbered,noend]{algorithm2e}
\usepackage[dvipsnames]{xcolor}
\usepackage[hidelinks]{hyperref}

\DeclareMathOperator*{\argmin}{arg\,min}
\newcommand{\SE}{\mathrm{SE}}
\newcommand{\R}{\mathbb{R}}
\newcommand{\Bset}{\mathcal{B}}
\newcommand{\Cset}{\mathcal{C}}
\newcommand{\Eset}{\mathcal{E}}
\newcommand{\Fset}{\mathcal{F}}
\newcommand{\Gset}{\mathcal{G}}
\newcommand{\Mset}{\mathcal{M}}
\newcommand{\Oset}{\mathcal{O}}
\newcommand{\Pset}{\mathcal{P}}
\newcommand{\Wset}{\mathcal{W}}

\title{\LARGE \bf
Backward Layout Search for Sequence-Constrained Robotic Assembly
}

\author{
Xi Zhang$^{1,*}$, 
Jiancong Dai$^{1,*}$, 
Hao Chen$^{1,\dagger}$,
Zhengtao Hu$^{2}$,
Changcai Yang$^{1}$, and Weiwei Wan$^{3}$%
\thanks{$^{1}$College of Computer and Information Sciences,
Fujian Agriculture and Forestry University, Fuzhou 350002, China.}%
\thanks{$^{2}$School of Mechatronic Engineering and
Automation, Shanghai University, Shanghai, China.}%
\thanks{$^{3}$Department of System Innovation,
Graduate School of Engineering Science, Osaka University, Toyonaka, Osaka, Japan.}%
\thanks{$^{*}$Xi Zhang and Jiancong Dai contributed equally.}%
\thanks{$^{\dagger}$Corresponding author: Hao Chen.
{\tt\small chenhaox@outlook.com.}}%
}

\begin{document}

\maketitle
\thispagestyle{empty}
\pagestyle{empty}

\begin{abstract} 
Robotic assembly layout planning must determine the assembly site and the initial pose of each part while ensuring collision-free execution of a prescribed assembly sequence. This problem is challenging because the obstacle environment changes after each assembly step, and unassembled parts remaining in the workspace may block robot motions. We observe that the feasibility of each assembly step depends only on the initial poses of the current and later-assembled parts. Based on this dependency, we propose Backward Layout Search (BLS), which assigns initial part poses in reverse assembly order. Each expansion performs geometric, kinematic, grasp, and prescribed-motion checks, while collision masks and candidate-set filtering remove infeasible initial part pose candidates. Promising partial layouts are retained through beam selection, and complete layouts are validated by full motion planning in forward assembly order. Experiments on five assembly models show that BLS produces collision-free executable layouts and reduces step evaluations and search time compared with a matched forward search. 
\end{abstract}

\section{INTRODUCTION}

Robotic assembly planning must consider both robot motion and workcell layout.
Research on assembly sequence planning has focused on determining how a
product should be assembled
\cite{tian2022assemble,tian2024asap,nagpal2024orasp,chen2021planning}, but has
paid less attention to where unassembled parts should initially be placed.
These initial poses directly affect assembly feasibility and motion cost
because the robot must pick, transfer, and assemble each part. This paper
addresses the following problem: given a prescribed assembly order and a set
of candidate grasps, select the assembly site and the initial pose of each part
such that the complete sequential assembly motion is collision-free and has a low
motion cost.


This problem is challenging because the obstacle environment evolves during
assembly. Before each pick, the obstacles include the fixed workcell, the
partial assembly, and all remaining unassembled parts at their initial poses.
Picking the current part frees the space at its initial pose; after insertion,
the part joins the partial assembly. Therefore, each assembly step has a
different obstacle environment, and the initial pose of a later part may block
several earlier motions.


\begin{figure}[!t]
\centering

\includegraphics[
    width=0.9\columnwidth
]{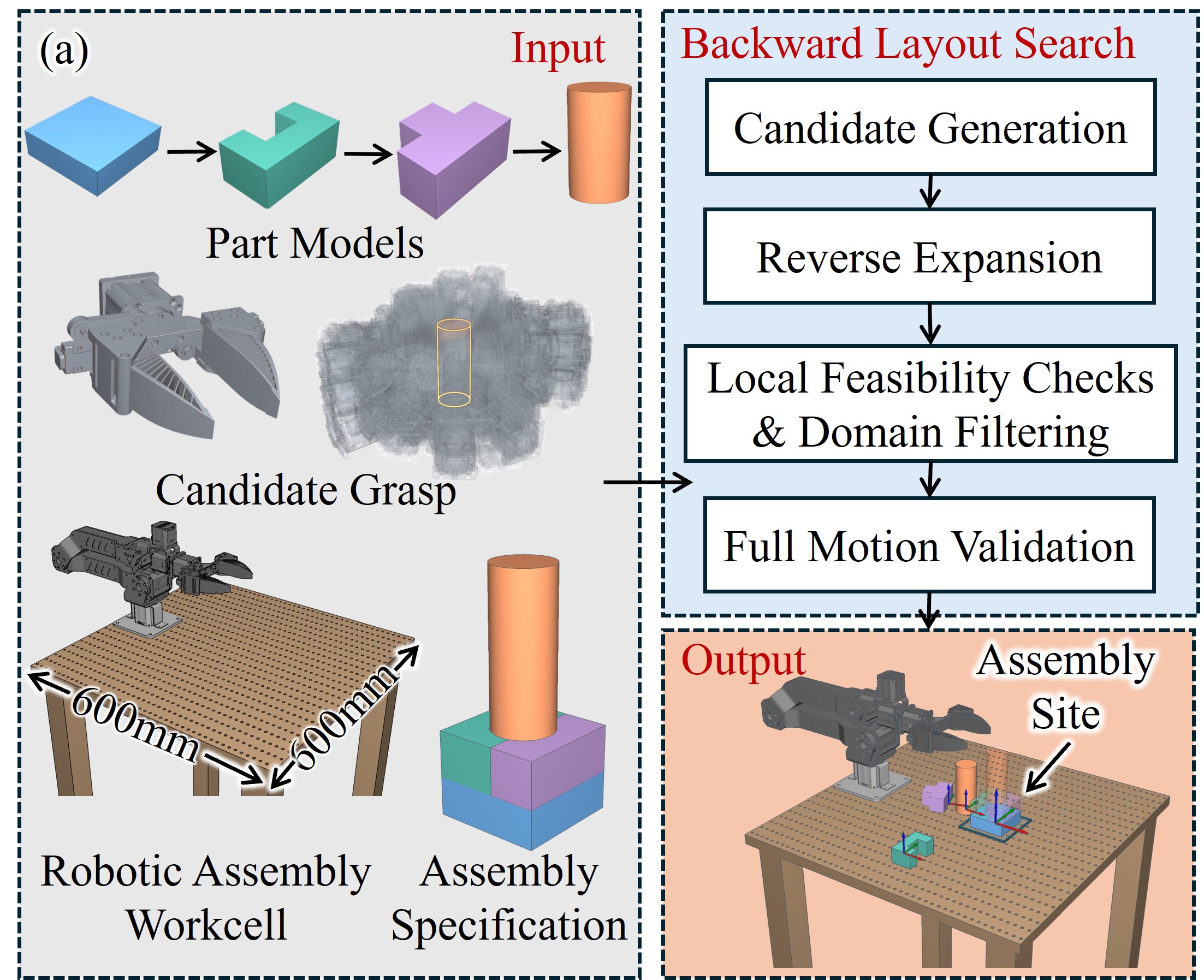}

\includegraphics[
    width=0.89\columnwidth
]{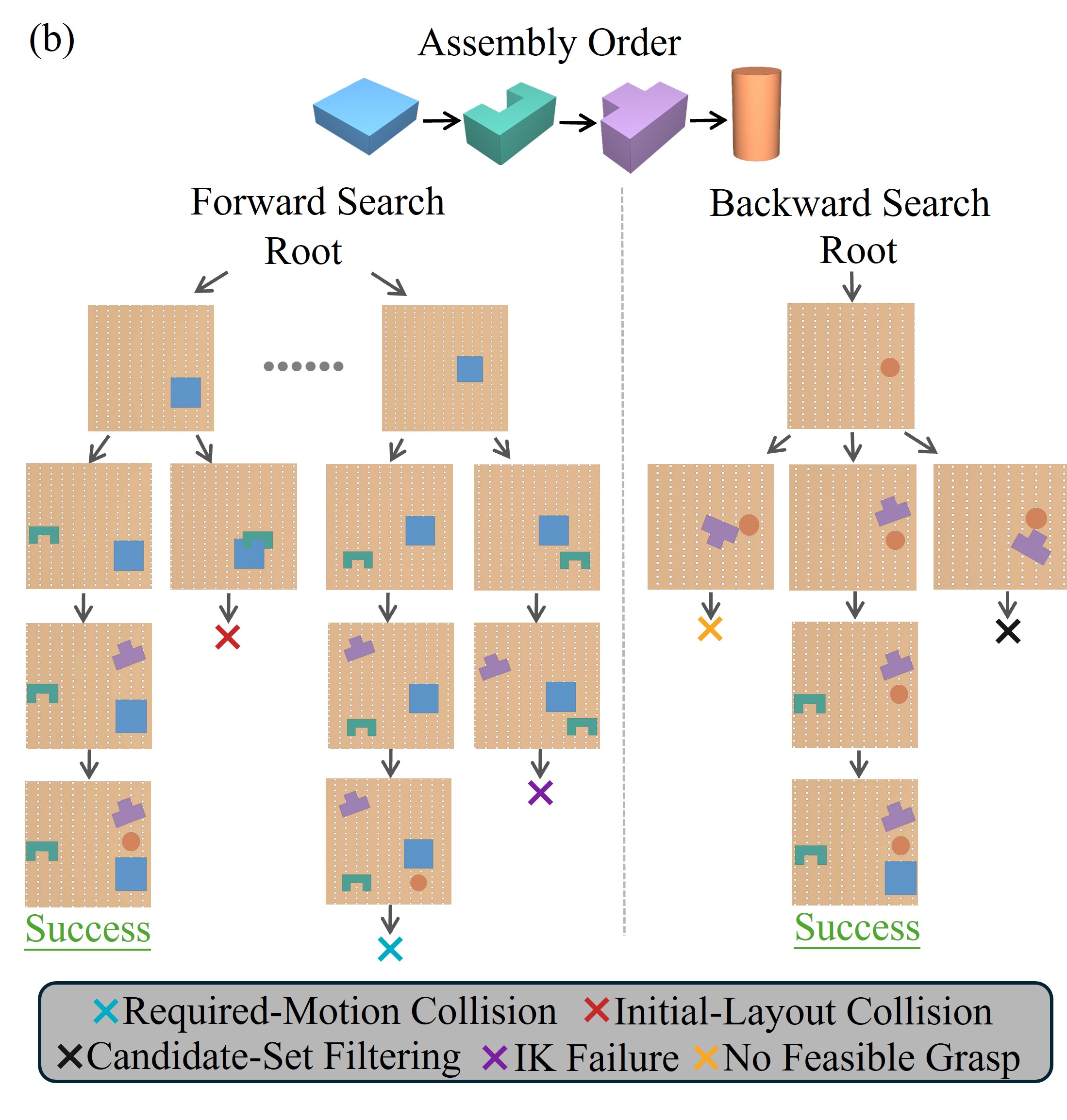}

\caption{Overview of the proposed Backward Layout Search (BLS).
(a) BLS planning pipeline.
(b) Illustration of forward and backward layout search using top-view. Part colors correspond to the assembly order shown above.}
\label{fig:bls_overview}
\end{figure}

Prior work on robotic workcell layout optimization has considered prescribed
task sequences, reachability, collision avoidance, travel distance, and
manipulability \cite{bachmann2021workcell}, as well as product information and
grasp-aware constraints \cite{beck2023assemblycell,ali2025taskgrasp}.
Motion-aware methods further couple layout decisions with trajectory, energy,
or execution-time objectives
\cite{qiu2022concurrent,kawabe2022motionlayout,baumgartner2025beyond}, and
surrogate models have been introduced to reduce repeated motion-planning cost
\cite{kawabe2025surrogate}. 
However, these methods generally evaluate each complete layout as a single
candidate and do not explicitly exploit the assembly-order structure of the
evolving obstacle environment. Consequently, an incompatible placement of a
later-assembled part may be detected only after many earlier poses have been
assigned, and similar layouts may require repeated inverse-kinematics,
collision, and motion-planning evaluations.

To address these limitations, we propose Backward Layout Search (BLS), as
illustrated in Fig.~\ref{fig:bls_overview}. For a fixed assembly site, the
execution feasibility of step \(k\) depends only on the initial poses of part
\(k\) and the later-assembled parts. BLS therefore assigns poses in reverse
assembly order, so each assignment determines one new step without changing
the later steps already checked. During expansion, local feasibility checks
reject infeasible candidates, while required-motion collision masks and
candidate-set filtering remove incompatible initial-pose candidates. Beam
selection retains the partial layouts with the lowest scores based on nominal
geometric costs. Complete layouts are then validated in forward order by full
motion planning. 

The main contributions of this paper are as follows: 
\begin{itemize}
    \item We formulate initial part layout planning for sequence-constrained robotic assembly. The formulation explicitly models the evolving obstacle environment and identifies the assembly-order dependency of step feasibility. 
    \item We propose BLS, which combines reverse-order assignment, local feasibility checks, required-motion collision masks, candidate-set filtering, and beam selection to efficiently search the discrete layout space. 
\end{itemize}

\section{RELATED WORK}
\subsection{Task-Aware Robotic Workcell Layout Optimization}

Task-aware workcell layout optimization extends geometric arrangement
by incorporating task requirements into layout evaluation.
Bachmann et al.~\cite{bachmann2021workcell} optimized component
placements for prescribed task sequences using waypoint reachability,
collision constraints, inter-task distance, and manipulability.
Other studies considered task positioning in multi-robot
cells~\cite{mutti2021taskpositioning} and workcell design based on
hierarchical manipulability maximization~\cite{franceschi2022workcell}.
Beck~\cite{beck2023assemblycell} combined layout and product-model assembly
information for automated 3D assembly-cell generation and validation.
Other studies integrated optimization, simulation, and digital-twin
technologies for reconfigurable and dynamic layout
design~\cite{arnarson2023smartlayout,wang2025dynamiclayout}, learning-based
online layout planning and scheduling~\cite{kaven2024multilayout}, and 3D
workcell layout planning under grasp and motion
constraints~\cite{ali2025taskgrasp}.

These studies establish task feasibility as a key layout criterion, but they
typically evaluate the complete workcell configuration as a single candidate.

\subsection{Motion-Aware Layout Optimization}
Motion-aware methods incorporate robot motion feasibility and execution cost
into layout optimization.
Qiu et al.~\cite{qiu2022concurrent} jointly optimized workcell layout
and robot trajectories with collision and energy objectives.
Kawabe et al.~\cite{kawabe2022motionlayout} coupled motion planning with
layout design in robotic cellular manufacturing systems.
They later introduced surrogate-assisted optimization to reduce the
computational cost of simultaneous motion and layout
design~\cite{kawabe2025surrogate}.
Baumg{\"a}rtner et al.~\cite{baumgartner2025beyond} showed that
geometric and joint-motion proxies may not accurately represent cycle
time and therefore evaluated layouts using time-optimal trajectory
duration.
Trajectory planning has also been incorporated into robot
base-placement optimization for industrial pick-and-place
sequences~\cite{wachter2024baseplacement}.

These methods model robot execution more directly, but evaluating many layout
candidates requires repeated inverse kinematics, collision checking, and
motion planning.

\subsection{Assembly Sequence and Motion Planning}

Assembly planning complements layout optimization by explicitly
reasoning about precedence constraints, partial assemblies, and
intermediate states.
Tian et al.~\cite{tian2022assemble} used disassembly reasoning to
generate assembly plans, illustrating backward reasoning for physical
assembly constraints.
Their subsequent ASAP framework~\cite{tian2024asap} searches physically
feasible assembly sequences while accounting for gravity and
partial-assembly stability.
Nagpal and Mehr~\cite{nagpal2024orasp} formulated assembly sequence
planning as a sequential decision-making problem and used precedence
constraints to prune the search space.
Kiyokawa et al.~\cite{kiyokawa2025many} combined simulation-based feasibility
and stability constraints with many-objective optimization for semi-automated
disassembly sequence planning.
Chen et al.~\cite{chen2021planning} planned block assembly with unstable
intermediate states using two manipulators, computing assembly orders,
candidate grasps, assembly directions, and supporting grasps.

These methods mainly optimize sequences and manipulation choices rather than
the initial spatial configuration of the parts. The interaction among initial part
poses, a prescribed assembly sequence, and the evolving obstacle
environment remains comparatively underexplored.

\section{PROBLEM FORMULATION}
\label{sec:problem}
\subsection{Assembly Model and Layout Variables}
We consider \(n\) rigid parts assembled by one robot in a prescribed order.
We index the parts by this order, so part \(k\) is assembled at step \(k\).
We assume direct manipulation: each part is transferred from its initial pose
to its assembly pose in a single grasp.

Let \(\Mset_i\) denote the geometry of part \(i\) in its body frame.
Let \(\Eset\) denote the fixed workcell environment, including the table and
the robot base.
Let \(\Cset\) be a finite set of candidate assembly sites. A candidate
\(a\in\Cset\) defines the world pose \(T_A(a)\in\SE(3)\) of the product frame.

The product model provides the final pose of part \(i\) relative to the
product frame, denoted by \({}^{A}T_i^f\in\SE(3)\).
The final world pose of part \(i\) is
\begin{equation}
T_i^f(a)=T_A(a)\,{}^{A}T_i^f.
\label{eq:final-pose}
\end{equation}
For each part, the assembly model also specifies an insertion motion that ends
at this final pose.

For each part \(i\), let \(\Pset_i\) be the finite set of candidate
initial positions. Let \(\Sigma_i\) be the set of stable support orientations, computed by the method in \cite{wan2015reorientating}. 
Let \(\Theta_i\) be the set of sampled in-plane rotations.
The initial layout decision for part \(i\)
is
\begin{equation}
x_i=(p_i,\sigma_i,\theta_i)\in\mathcal X_i,
\qquad
\mathcal X_i
=
\Pset_i\times\Sigma_i\times\Theta_i,
\label{eq:initial-placement}
\end{equation}
where \(p_i\in\Pset_i\) is an initial position. The support orientation \(\sigma_i\in\Sigma_i\) 
determines which part face contacts the table, and
\(\theta_i\in\Theta_i\) rotates the supported part about the table normal.
Together, \(x_i\) defines the initial part pose \(T_i^0(x_i)\in\SE(3)\).

Each part has a finite set of object-relative candidate grasps
\(\Gset_i=\{g_{i,1},\ldots,g_{i,N_i^g}\}\).
Each grasp specifies a fixed object-to-end-effector transform.
A grasp is not an independent layout variable: for each initial layout candidate, the planner
selects a grasp from \(\Gset_i\) and rejects the candidate if no grasp
satisfies the required kinematic, collision, and motion constraints.

We use $x_{k:n}=(x_k,\ldots,x_n)$
to denote the suffix of initial layout decisions from part \(k\) to part \(n\).

\subsection{Evolving Assembly State}
After steps \(1,\ldots,k\), the geometry of the partial assembly is
\begin{equation}
\Bset_k(a)
= 
\bigcup_{i=1}^{k}
T_i^f(a)\Mset_i.
\label{eq:partial-assembly}
\end{equation}
We define \(\Bset_0(a)=\varnothing\). The obstacle geometry changes within each assembly step. Before part \(k\) is picked, the obstacle set is
\begin{equation}
\Oset_k^{\mathrm{init}}
=
\Eset
\cup
\Bset_{k-1}(a)
\cup
\bigcup_{j=k}^{n}
T_j^0(x_j)\Mset_j.
\label{eq:initial-obstacles}
\end{equation}

After the pick, part \(k\) is attached to the end effector and included in the
robot collision model during transfer and insertion instead of acting as a
fixed obstacle. After release, it becomes part of the assembly. The
retreat obstacle set is
\begin{equation}
\Oset_k^{\mathrm{post}}
=
\Eset
\cup
\Bset_k(a)
\cup
\bigcup_{j=k+1}^{n}
T_j^0(x_j)\Mset_j.
\label{eq:post-obstacles}
\end{equation} 

\subsection{Local and Full Step Feasibility}

We first define a local feasibility check for efficient pruning.
It applies deterministic geometric and kinematic tests without sampling-based
path planning.

For step \(k\), let
\begin{equation}
\Gset_k^{\mathrm{loc}}(a,x_{k:n})
\subseteq
\Gset_k
\end{equation}
denote the grasps that satisfy all local conditions.
A grasp \(g\in\Gset_k\) belongs to
\(\Gset_k^{\mathrm{loc}}(a,x_{k:n})\) if it admits collision-free IK solutions at both
the initial and final part poses. The same grasp must also allow a
collision-free post-grasp retreat, prescribed insertion, and post-release
retreat.

The local feasibility indicator is
\begin{equation}
\psi_k(a,x_{k:n})
=
\begin{cases}
1,
&
\Gset_k^{\mathrm{loc}}(a,x_{k:n})
\neq\varnothing,\\
0,
&
\text{otherwise}.
\end{cases}
\label{eq:local-feasibility}
\end{equation}
The local check is necessary but not sufficient for complete
execution. It does not verify the transfer path from the initial pose to the pre-insertion
pose.

Let \(q_H\) be a fixed collision-free home configuration. Full step
feasibility requires a complete robot motion sequence:
\[
\begin{aligned}
q_H &\rightarrow \text{pick} \rightarrow \text{post-grasp retreat}
     \rightarrow \text{transfer} \\
    &\rightarrow \text{insertion} \rightarrow \text{release}
     \rightarrow \text{post-release retreat} \rightarrow q_H .
\end{aligned}
\]

Define \(\phi_k\) to indicate whether there exist a grasp
\(g\in\Gset_k^{\mathrm{loc}}(a,x_{k:n})\) and collision-free trajectories for
all segments of this sequence:
\begin{equation}
\phi_k(a,x_{k:n})
=
\begin{cases}
1,
& \text{if such a grasp and trajectories exist},\\
0,
&
\text{otherwise}.
\end{cases}
\label{eq:full-step-feasibility}
\end{equation}
Every fully feasible step must pass the local check:
\begin{equation}
\phi_k(a,x_{k:n})
\leq
\psi_k(a,x_{k:n}).
\label{eq:local-necessary}
\end{equation}

Let \(\Gamma(x_{1:n})\in\{0,1\}\) denote initial-layout feasibility.
The value \(\Gamma(x_{1:n})=1\) requires every initial part pose to lie
inside the tabletop region, avoid collision with \(\Eset\), and be
collision-free with all other initial parts.
A complete layout is feasible if
\begin{equation}
\Phi(a,x_{1:n})
=
\Gamma(x_{1:n})
\land
\bigwedge_{k=1}^{n}
\phi_k(a,x_{k:n})
=
1.
\label{eq:complete-feasibility}
\end{equation}

\subsection{Assembly-Order Dependency}



During forward execution of step \(k\), parts \(1,\ldots,k-1\) no longer
occupy their initial poses and appear only in the fixed partial assembly
\(\Bset_{k-1}(a)\). In contrast, parts \(k,\ldots,n\) remain at their initial
poses before the pick. Therefore, \(\psi_k\) and \(\phi_k\) depend on the
assembly site \(a\) and suffix \(x_{k:n}\), but not on
\(x_{1:k-1}\). This yields the safe pruning rule
\begin{equation}
\psi_k(a,x_{k:n})=0
\Longrightarrow
\Phi(a,x_{1:n})=0,
\quad
\forall x_{1:k-1}.
\label{eq:suffix-pruning}
\end{equation}
Hence, an infeasible suffix can be discarded before earlier poses are
generated. Moreover, once \(x_{k:n}\) has been evaluated, assigning
\(x_{k-1}\) does not require reevaluating steps \(k,\ldots,n\), which motivates
the reverse search order.

\subsection{Optimization Objective}
The layout search problem treats full feasibility as a hard constraint.
During BLS, local feasibility checks provide necessary pruning conditions,
and full feasibility is evaluated after a complete layout is generated.

Let \(\mathbf{p}_H\in\R^3\) be the tool-center point at \(q_H\). Let
\(\mathbf{p}_i^0(x_i)\) and \(\mathbf{p}_i^f(a)\) be the world positions of
the same part-fixed reference point at the initial and final poses,
respectively.

Let \(\bar{\mathbf{p}}_i^0(x_i)\) and
\(\bar{\mathbf{p}}_i^f(a)\) be fixed-height points above
these two poses. The nominal geometric cost of part \(i\) is
\begin{equation}
\begin{split}
c_i^{\mathrm{nom}}(a,x_i)
={}&
\|\mathbf{p}_H-\bar{\mathbf{p}}_i^0\|_2
+
\|\bar{\mathbf{p}}_i^0-\mathbf{p}_i^0\|_2\\
&+
\|\bar{\mathbf{p}}_i^0-\bar{\mathbf{p}}_i^f\|_2
+
\|\bar{\mathbf{p}}_i^f-\mathbf{p}_i^f\|_2\\
&+
\|\bar{\mathbf{p}}_i^f-\mathbf{p}_H\|_2.
\end{split}
\label{eq:nominal-cost}
\end{equation}
The five terms respectively approximate the home-to-approach, pickup,
transfer, insertion, and return segments. This cost is deterministic and inexpensive to evaluate but does not represent the final joint-space trajectory length. The total nominal cost is
\begin{equation}
J_{\mathrm{nom}}(a,x_{1:n})
=
\sum_{i=1}^{n}
c_i^{\mathrm{nom}}(a,x_i).
\label{eq:total-nominal-cost}
\end{equation}
The layout problem is
\begin{equation}
\begin{aligned}
(a^\star,x_{1:n}^\star)
=
\argmin_{a,x_{1:n}}
&\quad
J_{\mathrm{nom}}(a,x_{1:n})\\
\mathrm{s.t.}
&\quad
a\in\Cset,\\
&\quad
x_i\in\mathcal X_i,
\quad i=1,\ldots,n,\\
&\quad
\Gamma(x_{1:n})=1,\\
&\quad
\phi_k(a,x_{k:n})=1,
\quad k=1,\ldots,n.
\end{aligned}
\label{eq:layout-optimization}
\end{equation}

\section{BACKWARD LAYOUT SEARCH}
\label{sec:method}

\subsection{Overview}

Backward Layout Search (BLS) assigns initial poses in reverse assembly order,
$(x_n,x_{n-1},\ldots,x_1)$.
At reverse depth \(k\), node \(N_k\) stores: 1) the assigned suffix \(x_{k:n}\); 2) its
feasible grasps; 3) the candidate sets \(\mathcal X_i(N_k)\) for unassigned
parts \(i<k\); and 4) the accumulated nominal cost.
The root \(N_{n+1}\) has no assigned pose and is first expanded by
assigning \(x_n\).







\subsection{Candidate Set Generation}

The allowed assembly region is discretized into a finite set of assembly sites
\(\Cset\). All optimality statements therefore refer to this discrete domain.
For each site \(a\), the initial-pose sets \(\mathcal X_i\) in
Eq.~\eqref{eq:initial-placement} are filtered by tabletop and environment
collision constraints, robot-base clearance, initial-pose IK, insertion, and
post-release retreat feasibility.
Each part is checked independently at this stage without other unassembled
parts in the scene. Inter-part constraints are introduced during reverse
expansion.

The surviving poses form the root candidate sets
\begin{equation}
\mathcal X_i(N_{n+1})\subseteq\mathcal X_i.
\label{eq:root-candidate-set}
\end{equation}
The assembly site is rejected if any root candidate set is empty.

\subsection{Reverse Expansion and Local Feasibility Checks}

Assume that node \(N_{k+1}\) contains the assigned suffix
\(x_{k+1:n}\). BLS expands this node by selecting \(x_k\in\mathcal X_k(N_{k+1})\),
which forms the new suffix \(x_{k:n}\). BLS then computes
the locally feasible grasp set
\(\Gset_k^{\mathrm{loc}}(a,x_{k:n})\) defined in
Section~\ref{sec:problem}. 

The local check includes initial-layout collision, endpoint IK, common
grasp feasibility, post-grasp retreat, insertion, and post-release
retreat. The child is rejected if
\(\Gset_k^{\mathrm{loc}}(a,x_{k:n})=\varnothing\).
Sampling-based path planning is not used for partial-node pruning.
It is applied only during the final validation of complete initial
layouts.

\subsection{Required-Motion Collision Masks}

For a candidate initial placement \(x_k\), assembly site \(a\), and
grasp \(g\), let
\begin{equation}
\Wset_k^{\mathrm{req}}(a,x_k,g)
\end{equation}
denote the swept volume of the required local motions.



The swept volume covers the post-grasp retreat, prescribed insertion, and
post-release retreat motions.

For a later part \(j>k\), define the forbidden initial-pose set
\begin{equation}
\Fset_{k,j}(a,x_k,g)
=
\left\{
x_j\in\mathcal X_j
\;\middle|\;
T_j^0(x_j)\Mset_j
\cap
\Wset_k^{\mathrm{req}}(a,x_k,g)
\neq\varnothing
\right\}. 
\label{eq:forbidden-set}
\end{equation}
If the assigned decision \(x_j\in\Fset_{k,j}(a,x_k,g)\),
then grasp \(g\) cannot execute the required local motions for step
\(k\) and is removed. If no grasp remains,
the candidate \(x_k\) is rejected. 

After assigning \(x_k\), the child initializes
\(\mathcal X_i(N_k)\) from the parent sets
\(\mathcal X_i(N_{k+1})\) for all \(i<k\). BLS filters the candidate sets of all earlier parts
using initial-pose collision, tabletop, local-feasibility, and grasp
constraints. The child is rejected if
\(\mathcal X_i(N_k)=\varnothing\) for any \(i<k\). 

\subsection{Beam Selection}
For node \(N_k\), the beam score combines the nominal cost of the assigned
suffix with an optimistic completion estimate for the unassigned parts:
\begin{equation}
S(N_k)
=
\sum_{j=k}^{n}
c_j^{\mathrm{nom}}(a,x_j)
+
\sum_{i=1}^{k-1}
\min_{x_i\in\mathcal X_i(N_k)}
c_i^{\mathrm{nom}}(a,x_i).
\label{eq:beam-score}
\end{equation}
The first term is the nominal cost of the assigned suffix. For each
unassigned part, the second term selects the least-cost pose remaining in its
current candidate set. The estimate
ignores interactions among those parts and is used only for beam
ranking. At the complete depth
\(k=1\), no unassigned part remains, so
\(S(N_1)=J_{\mathrm{nom}}(a,x_{1:n})\).

At each reverse depth \(k\), BLS scores all surviving children and uses the
\(B\) lowest-scoring nodes as the next frontier. A smaller \(B\) reduces
subsequent expansions and evaluations, whereas a larger \(B\) retains more
alternative suffixes. 
Beam pruning makes BLS incomplete and therefore does not guarantee a global
optimum over the discrete candidate sets. Regardless of its beam score, every
reported layout must pass full motion validation.

\subsection{Full Motion Validation}

Sampling-based path planning is used only at this stage, not for partial-node
pruning. Surviving complete layouts are validated in the prescribed forward
assembly order using the full motion planner defined in
Section~\ref{sec:problem}. The workcell state is updated after each successful
step, and only layouts whose complete sequence succeeds are accepted. A
planning timeout is treated as failure under the assigned budget but does not
establish mathematical infeasibility.

\begin{algorithm}[t]
\caption{Backward Layout Search}
\label{alg:bls}
\KwIn{assembly sites \(\Cset\), initial-pose sets \(\mathcal X_i\), grasp sets
\(\Gset_i\), beam width \(B\)}
\KwOut{validated pair \((a,x_{1:n})\) with the lowest
\(J_{\mathrm{nom}}\), or failure}

\(\textit{best}\leftarrow\varnothing\),
\(J_{\mathrm{nom}}(\varnothing)\leftarrow\infty\)\;
\ForEach{\(a\in\Cset\)}{
    Build the root candidate sets \(\mathcal X_i(N_{n+1})\), \(i=1,\ldots,n\)\;
    \lIf{\(\exists i:\mathcal X_i(N_{n+1})=\varnothing\)}{\textbf{continue}}
    \(\textit{frontier}\leftarrow\{N_{n+1}\}\)\;
    \For{\(k=n,n-1,\ldots,1\)}{
        \(\mathcal N_k\leftarrow\varnothing\)\;
        \ForEach{\(N_{k+1}\in\textit{frontier}\) and
        \(x_k\in\mathcal X_k(N_{k+1})\)}{
            Compute \(\Gset_k^{\mathrm{loc}}(a,x_{k:n})\) by the local checks
            and the required-motion masks\;
            \lIf{\(\Gset_k^{\mathrm{loc}}(a,x_{k:n})=\varnothing\)}{\textbf{continue}}
            Create child \(N_k\) and filter \(\mathcal X_i(N_k)\) from
            \(\mathcal X_i(N_{k+1})\) for all \(i<k\)\;
            \lIf{\(\exists i<k:\mathcal X_i(N_k)=\varnothing\)}{\textbf{continue}}
            \(\mathcal N_k\leftarrow\mathcal N_k\cup\{N_k\}\) with score
            \(S(N_k)\)\;
        }
        \(\textit{frontier}\leftarrow\) at most \(B\) lowest-scoring nodes of
        \(\mathcal N_k\)\;
        \lIf{\(\textit{frontier}=\varnothing\)}{\textbf{break}}
    }
    \ForEach{complete node \(N_1\in\textit{frontier}\)}{
        Validate \((a,x_{1:n})\) in forward assembly order\;
        \lIf{validation succeeds and
        \(S(N_1)<J_{\mathrm{nom}}(\textit{best})\)}{\(\textit{best}\leftarrow(a,x_{1:n})\)}
    }
}
\lIf{\(\textit{best}=\varnothing\)}{\Return failure}
\Return \(\textit{best}\)\;
\end{algorithm}

\section{EXPERIMENTS}
\label{sec:experiments}

\subsection{Experimental Platform and Assembly Models}

\begin{figure}[t]
\centering
\includegraphics[width=0.9\columnwidth]{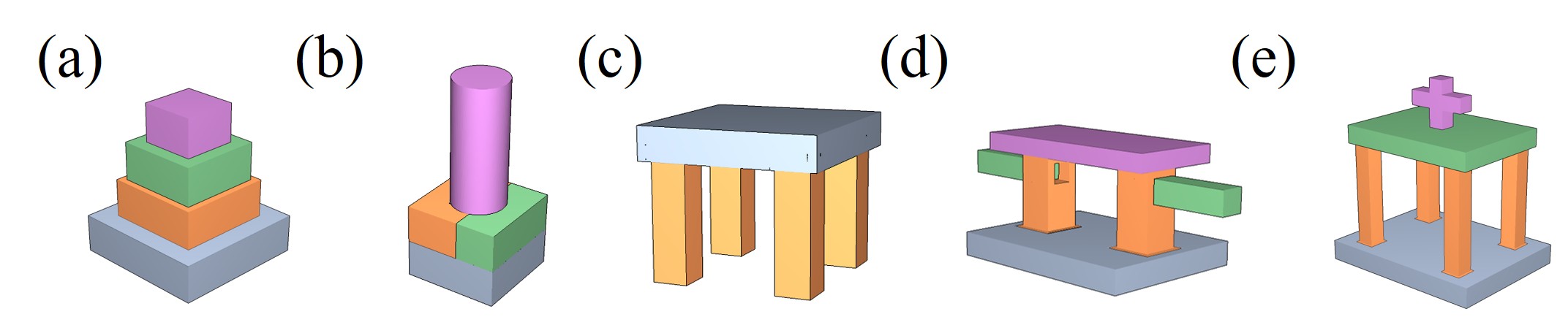}
\caption{Assembly models used in the experiments:
(a) pyramid, (b) hammer, (c) chair,
(d) frame, and (e) tower.}
\label{fig:assembly_geometries}
\end{figure}

All experiments use our custom-designed six-degree-of-freedom manipulator and
run on a workstation equipped with an AMD Ryzen 7 9800X3D CPU. 
The implementation is in Python and uses CPU computation only.
 Fig.~\ref{fig:assembly_geometries} shows the pyramid, hammer, chair, frame, and
tower models, which provide different part geometries, support orientations,
assembly structures, and evolving obstacle conditions. 

Every compared layout is evaluated with the same rule: a collision check
at the home configuration \(q_H\), the local feasibility checks of
Section~\ref{sec:problem}, and full motion validation. For a layout that
passes validation, we report the TCP travel that is the Cartesian path
length of the TCP over the complete validated assembly sequence.
\textbf{The attached video shows the execution of the returned layouts on the real
robot.}

\subsection{BLS Initial Layouts and Spatial Interpretation}

\begin{figure}[t]
\centering
\includegraphics[width=0.9\columnwidth]{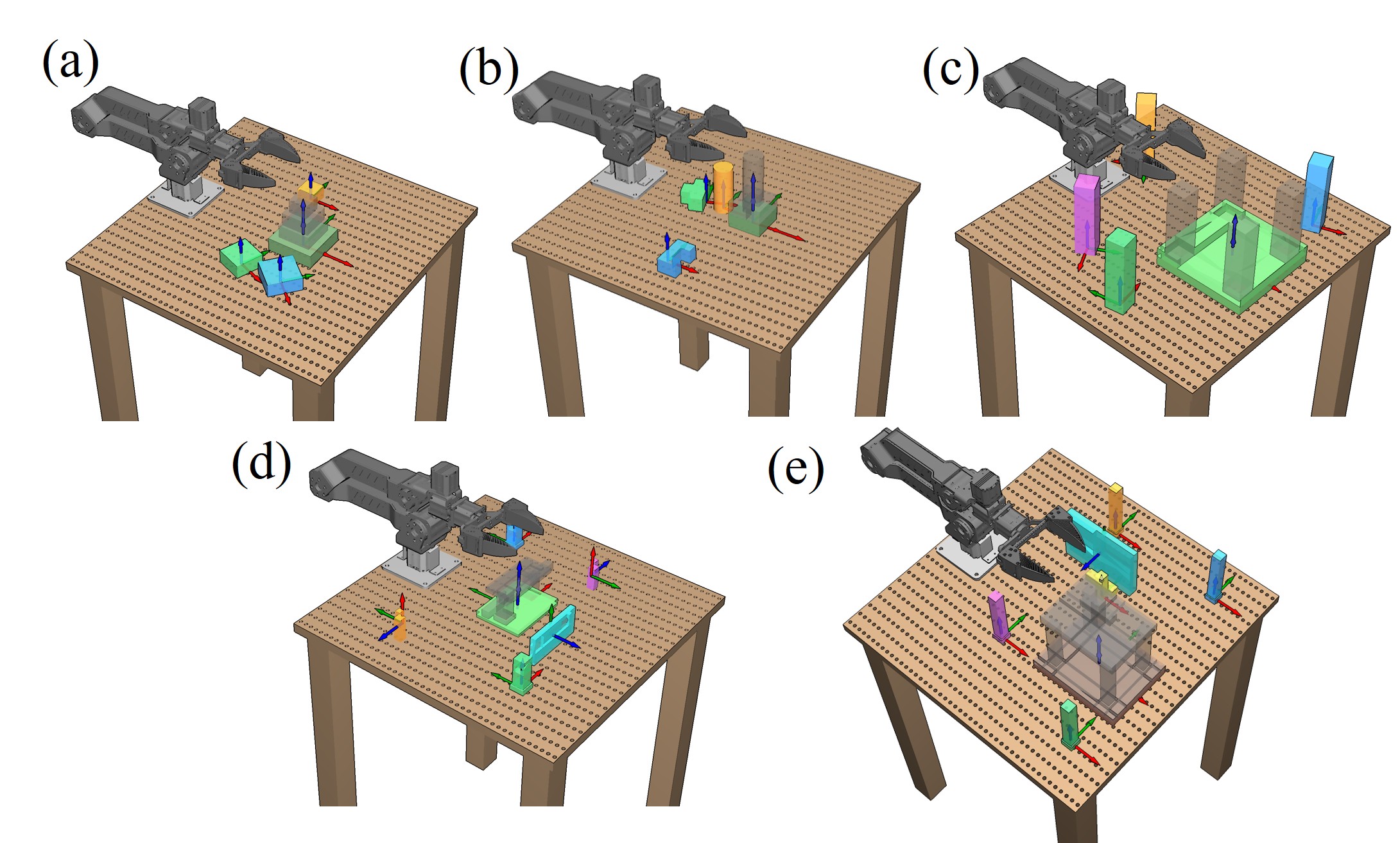}
\caption{Initial layouts and assembly sites returned by BLS for the five
assembly models: (a) pyramid, (b) hammer, (c) chair, (d) frame, and
(e) tower. Solid parts denote the initial part poses, whereas translucent
parts visualize the eventual final assembly at the selected assembly site.}
\label{fig:product_model_layouts}
\end{figure}

Fig.~\ref{fig:product_model_layouts} shows the assembly sites and initial
layouts returned by BLS. Several initial parts lie within the planar footprint of the translucent final
assembly. This does not indicate a collision because the translucent geometry
only visualizes the final product and is not a static obstacle. 
In every returned layout,
the first part is placed directly at its final
assembly pose. This placement is admissible because the partial assembly is
still empty at step one, so the pose cannot block an earlier motion.

\subsection{Backward Versus Forward Search}

To assess the effect of initial-pose assignment order, BLS is compared with
an iterative forward search, denoted Forward-Iter. Both methods use the same
assembly-site set \(\Cset\), initial-pose candidate sets \(\mathcal X_i\),
grasp sets \(\Gset_i\), collision models, nominal geometric cost
\(J_{\mathrm{nom}}\), local feasibility checks, and full motion validation
procedure.BLS assigns poses in reverse assembly order and reuses suffix
feasibility results, whereas Forward-Iter assigns poses in forward order and
repeatedly evaluates the growing prefix. 

As shown in Table~\ref{tab:search_order}, BLS has lower search time for
all five assembly models and requires fewer step evaluations in four of the
five cases. The chair is the only exception in evaluation count, although BLS
remains faster. Step evaluations count step-level feasibility checks rather
than fixed-cost operations, so the runtime of each evaluation varies with the
grasp, IK, collision, required-motion, and candidate-set filtering operations
performed at each node. Therefore, search time need not scale linearly with the evaluation count. 
Nominal costs are comparable: BLS obtains lower values
for the chair, frame, and tower, the same value for the pyramid, and a slightly
higher value for the hammer.

\begin{table}[t]
\centering
\caption{Comparison of BLS and Forward-Iter}
\label{tab:search_order}
\scriptsize
\renewcommand{\arraystretch}{1.05}
\setlength{\tabcolsep}{2.5pt}
\resizebox{\columnwidth}{!}{%
\begin{tabular}{llrrr}
\toprule
Assembly & Method & Nominal cost & Step evaluations & Search time (s) \tabularnewline
\midrule

(a) Pyramid
& BLS
& \textbf{2.156}
& \textbf{336}
& \textbf{109.50} \tabularnewline
& Forward-Iter
& \textbf{2.156}
& 624
& 157.15 \tabularnewline

\midrule
(b) Hammer
& BLS
& 2.092
& \textbf{368}
& \textbf{104.49} \tabularnewline
& Forward-Iter
& \textbf{2.073}
& 608
& 139.96 \tabularnewline

\midrule
(c) Chair
& BLS
& \textbf{2.890}
& 432
& \textbf{77.01} \tabularnewline
& Forward-Iter
& 3.067
& \textbf{400}
& 98.02 \tabularnewline

\midrule
(d) Frame
& BLS
& \textbf{3.742}
& \textbf{208}
& \textbf{110.70} \tabularnewline
& Forward-Iter
& 4.018
& 336
& 140.00 \tabularnewline

\midrule
(e) Tower
& BLS
& \textbf{4.557}
& \textbf{480}
& \textbf{121.85} \tabularnewline
& Forward-Iter
& 4.964
& 816
& 204.41 \tabularnewline

\bottomrule
\end{tabular}%
}
\end{table}

\begin{table*}[!htbp]
\centering
\caption{Feasibility and validated TCP travel for the five assembly models.}
\label{tab:feasibility_compare}
\footnotesize
\renewcommand{\arraystretch}{1.05}
\setlength{\tabcolsep}{3.5pt}

\begin{tabularx}{\textwidth}{
p{2.7cm}
*{5}{>{\centering\arraybackslash}X}
}
\toprule
Method / Metric &
(a) Pyramid &
(b) Hammer &
(c) Chair &
(d) Frame &
(e) Tower \tabularnewline
\midrule

Bachmann et al.~\cite{bachmann2021workcell}
& Fail (no feasible grasp)
& Fail (no feasible grasp)
& Fail (home collision)
& Fail (home collision)
& Fail (home collision) \tabularnewline

TCP travel (m)
& --
& --
& --
& --
& -- \tabularnewline

\midrule

Ali et al.~\cite{ali2025taskgrasp}
& Pass
& Pass
& Fail (home collision)
& Fail (no feasible grasp)
& Fail (no feasible grasp) \tabularnewline

TCP travel (m)
& 1.714
& 2.069
& --
& --
& -- \tabularnewline

\midrule

\textbf{BLS (ours)}
& \textbf{Pass}
& \textbf{Pass}
& \textbf{Pass}
& \textbf{Pass}
& \textbf{Pass} \tabularnewline

TCP travel (m)
& \textbf{0.962}
& \textbf{1.610}
& 2.784
& 3.677
& 4.991 \tabularnewline

\bottomrule
\end{tabularx}
\end{table*}

\subsection{Feasibility Comparison with Prior Layout Methods}

We reproduce the layout methods of Bachmann
et al.~\cite{bachmann2021workcell} and Ali et al.~\cite{ali2025taskgrasp} in
the same workcell and on the same five assembly models.
Table~\ref{tab:feasibility_compare} reports the outcome of each reproduced
layout under the same evaluation rule. This comparison
characterizes the reproduced layouts under our execution requirements rather
than a general limitation of the original methods.

The reproduced Bachmann layouts fail on all five models. On the pyramid and
the hammer, the placed parts leave insufficient clearance for any candidate
grasp, whereas on the chair, frame, and tower the robot is already in
collision at \(q_H\), so no assembly step can start. The reproduced Ali
layouts pass on the pyramid and the hammer. They fail on the chair by home
collision and on the frame and tower by no feasible grasp.
BLS passes full motion validation on all five models and is therefore the only
compared method that produces an executable layout for the chair, frame, and
tower. TCP travel can be compared only where both methods return a validated
motion. In these two cases, BLS reduces it from 1.714\,m to 0.962\,m for the
pyramid and from 2.069\,m to 1.610\,m for the hammer.

\section{CONCLUSION}
This paper presented Backward Layout Search (BLS) for initial part layout
planning under a prescribed assembly order. BLS exploits the assembly-order
dependency of step feasibility to assign initial poses in reverse order, reuse
previously evaluated suffix feasibility, and prune infeasible suffixes before
earlier poses are generated. Candidate-set filtering and beam selection bound
the search effort, and every surviving complete layout is accepted only after
full motion validation.

Across five assembly models, BLS achieves lower search time than Forward-Iter
in all cases and fewer step evaluations in four of them, with comparable
nominal cost, and it passes full motion validation for all five models. Under
the same evaluation rule, the reproduced prior layouts fail in several cases because
of home-configuration collisions or insufficient grasp clearance.

\small
\bibliographystyle{IEEEtran}
\bibliography{references}

\end{document}